*Commentary*

# Imperfect Match: PDDL 2.1 and Real Applications

**Mark Boddy**  MARK.BODDY@ADVENTIUMLABS.ORG
*Adventium Labs*
*111 Third Avenue S.*
*Minneapolis, MN 55401 USA*

### Abstract

PDDL was originally conceived and constructed as a *lingua franca* for the International Planning Competition. PDDL2.1 embodies a set of extensions intended to support the expression of something closer to "real planning problems." This objective has only been partially achieved, due in large part to a deliberate focus on not moving too far from classical planning models and solution methods.

## 1. Introduction

Fox and Long (2003) describe a set of extensions to PDDL. They claim that with these extensions "PDDL2.1 begins to bridge the gap between basic research and applications-oriented planning by providing the expressive power necessary to capture real problems." While the expressive extensions embodied in PDDL2.1 represent a step in the right direction, the authors' claim is true only in a limited sense. The rest of this commentary will attempt to make the qualifying reservations clear.

## 2. Planning and Real Applications

There is no such thing as a "real planning problem," any more than there is such a thing as a "real mathematical programming problem." There are real problems, for which people have found it useful to construct models and apply solution techniques that are generally viewed as planning models and techniques (or math programming techniques, or something else entirely). Examples of problems for which planning models may be useful can be drawn from manufacturing (e.g., batch manufacturing operations), or from the high-level control of complex mechanisms such as spacecraft. For math programming, to continue the comparison, there are disciplines such as Chemical Engineering, a major branch of which is concerned with how to model and solve problems in chemical manufacturing using various flavors of mathematical optimization. There is no equivalent discipline for planning, and so it is more difficult to characterize "planning applications" in the same sense in which control of chemical processes may be described as an application for mathematical optimization.

This lack of an extensive history of applications makes it difficult to assess the relevance of particular modeling or solution techniques, including those developed within the planning research community. The International Planning Competition (IPC), held at AIPS-98, AIPS-00 and AIPS-02, has been attempting to provide the data necessary to make these assessments. Unfortunately, the approach taken gets things backwards, implicitly asking

"how can we extend the models everybody in a particular community is comfortable working with, so that they are closer to what we think is needed for a real application?"

If planning research is to be viewed as a discipline akin to mathematics, this kind of extension from known theoretical constructs makes sense. If planning is engineering and real applications are the point, theoretical work should emerge from, or at the very least be tested against, generalization and abstractions of real applications. The classical planning approach grew out of work on theorem-proving and dynamic logics, and ever since has most fruitfully been applied in domains involving minimal interaction with the physical world, which describes some planning domains (softbots and other software agent-based applications, for example), but not the vast majority of them. Of the systems cited in Fox and Long's Introduction as having been applied to real planning problems (SIPE, O-Plan, HSTS, and IxTeT), none are conventional classical planners constructing totally-ordered sequences of operators. All of them use some combination of methods drawn from temporal network planning and HTN (task decomposition) planning. Other work applying planning methods to real applications, for example ASPEN (Fukunaga et al., 1997), does not make much use of classical planning techniques, either. Perhaps classical planning is addressing the wrong problem, or the wrong parts of the problem, for most real applications.

## 3. Evaluating the Language Itself

As argued by a long line of people going back at least to Hendrix (1973), a point-based temporal model in which instantaneous events effect changes in the world state can be used to build very expressive models of system dynamics, simply by treating those events as the starting or ending points of intervals over which propositions hold or processes change the world state. Over the past decade or so, Reiter (2001) and others have extended the semantics of the Situation Calculus (McCarthy & Hayes, 1969) to encompass overlapping actions, metric quantities, continuous change, exogenous actions, limited knowledge, contingent action effects, and actions with uncertain effects, among other things. As presented by Fox and Long, PDDL2.1 can be used to express many of the things one might want to represent for a real problem. However, there is a difference between saying that something can be expressed in a given language, and that that expression is natural, intuitive, or easy to use.

Take the modeling of resources as an example: PDDL2.1 is expressive enough to represent unary resources (trucks, tools), capacity resources (power, weight), and consumable resources (fuel). But there is no easy way in the language to refer to properties of the resource itself. Everything is encoded in the action representation, as for example in the modeling of TOTAL-FUEL-USED in the example of Figure 5. In many real domains, resources are primary, in the sense that the hard part of the problem is figuring out how to resolve resource conflicts, within the other constraints in the problem statement.

### 3.1 Form Follows Function

The significance of the distinction between what can expressed and how easily it can be expressed or manipulated depends on what roles the language is intended to fill. PDDL2.1 is in its origins and current use a language for the planning competition. As long as the current language permits the (not necessarily terribly natural) expression of required features of the domain and planning problem, perhaps there is no problem. Competitors who wish to construct special-purpose data structures such as explicit representations of resources are free to do so.

If PDDL2.1 is to be employed beyond the IPC, what role(s) are intended? A high-level modeling language for real applications has very different requirements from a competition standard, different again from a language used to pass planning problems and solutions in machine-readable forms, different yet again from a modular, locally-extensible language used as a research tool, permitting researchers to exchange example domains so as to be able to compare results using different techniques on the same data.[1]

### 3.2 Specific Problems with the Language

Finally, there are some specific problems with the current semantics of PDDL2.1, which should be addressed pretty much whatever use it is to be put to.

#### 3.2.1 Holdovers From the Classical Model

Fox and Long are explicit about having made choices in both syntax and semantics to keep a familiar feel to the language for the classical planning community. This makes good sense if the objective is to use the language to push the boundaries of what can be done using or extending the models and methods currently popular in the research community, but does little to support, and may in some cases actively hinder, the introduction of features drawn from real applications.

There are several problems resulting from this stance. First, the language is explicitly restricted to ensure a finite set of ground actions, specifically because some planning algorithms require this. This precludes the modeling of a number of different kinds of domain features, for example the explicit creation and destruction of objects (e.g., intermediate data products, in an image-processing application), or flexible solution for continuous parameters (for example, trajectory optimization) within an existing or evolving plan. Shouldn't the models and methods be following the requirements of the problems to be solved, rather than the other way round?

A second problem is the treatment of durative actions as atomic, rather than treating activities like heating water or lighting matches as starting processes that may proceed on their own, until the agent decides again to intervene. Consider the HEAT-WATER action defined in Figure 12 of Section 5.3, whose duration is defined to be precisely that required to raise the water to 100 C. This has the curious result that in any well-defined plan, the duration of HEAT-WATER must be exactly the time required, taking into account any overlapping actions that may also affect the temperature of the water, where those actions may not be specified until after the HEAT-WATER action is added to the plan.

Finally, the authors define a semantics for durative actions in which preconditions are required to be true at the beginning or through the extent of, and postconditions are asserted at the end of, intervals of non-zero width open on the right. This is clean, clear, and requires no special constraints to have a well-defined semantics, at least for totally-ordered begin and end-points. AT-END preconditions mess this up, and furthermore are the wrong solution for what the authors appear to be trying to support, which I take to be the implicit expression of complex structure within a durative action ("this action requires p to be true for 2 minutes, from the start of the action"). See Dave Smith's commentary, also in this issue, for a discussion of what a more complete model of complex durative actions might look like, if one were less concerned with staying close to the classical planning model.

---

1. For a summary of previous attempts to construct a shared ontology for planning, see Tate (1998).

### 3.2.2 Problems with the Definition of the Continuous Model

In Section 5.3, Fox and Long introduce a notation for specifying durative actions with continuous effects, claiming that using their #t notation, it is possible to express a variety of different nonlinear functions of time. At least as presented in that section, the #t construction does not appear to permit the definition, implicit or otherwise, of nonlinear continuous functions. In their example of expressing the effect of acceleration on position by making velocity vary over time, evaluating position at #t by multiplying #t by a time-varying velocity also evaluated at #t will result in an incorrect answer. The required operation is integration, not composition.

Also in that section, the authors say that continuous durative actions don't support exogenous events. I don't understand why they can't, as long as the endpoints of those exogenous events also appear in the plan. Incorporating this capability would significantly increase the range of real (or realistic) problems that could be represented.

### 3.2.3 Other Issues

Two other matters are worth pointing out as well. First is the restriction to numeric domains for functions, for which a primary motivation cited is the previously-discussed objective of making it possible to construct finite extensional models of the set of possible actions. Nonnumeric functions are frequently useful in real domains, for example to refer to the current state of an object (the current configuration of a piece of communication equipment, say).

Finally, the authors' decision to support "Undefined" values for numeric fluents is by their statement intended to allow actions to determine fluent values by setting them, prior to which the value would be unknown. Representing incomplete information, including unknown propositional and continuous states, is an important capability, and one that should be addressed directly. It might deliberately be left as a later extension but I do not understand the merits of addressing it incompletely in this way.

## 4. Summary and Conclusions

The extensions embodied in PDDL2.1 are a definite step in the direction of a language in which complex planning problems can be expressed. Fox and Long have done the planning community a considerable service, first, in designing and implementing those extensions, and second, in presenting and motivating them in the current paper. The focus on backward compatibility with classical planning is problematic, given their expressed desire to address real (or at least more realistic) applications, and there are some other necessary cleanups to the language, noted above.

Any further extensions to PDDL2.1 will have to address the very likely fragmenting of the field into mutually incompatible sub-fields. The logical end of this process can be seen in the survey of scheduling problems compiled by Graham et al. (1977), in which minor differences in problem statement lead to very different complexity classes. Classification by complexity class may be less relevant in a field where almost every interesting problem is at least NP-hard, but a similar phenomenon will arise from the fact that some problems require different expressive capabilities than others (resource constrained project scheduling problems, versus manuver planning for satellite constellations, for example), or are hard to solve in very different ways (finding a correct ordering on plan steps, versus resolving conflicts on a unary resource).

Ultimately, techniques developed in the classical planning literature will probably prove useful in addressing complex, real-world applications. However, I do not believe that they will in most cases be central to those solutions. The classical planning focus is on individual actions, rather than organizing or synchronizing with operations in the larger environment, on discrete state changes, rather than multiple, interacting asynchronous processes, and on propositional representations, rather than constraints on continuous quantities. Other methods will have to be brought to bear to address the core complexities of most real "planning" problems, which involve complex resources and other forms of synchronized behavior, exogenous events, temporal uncertainties in both the agent's own actions and other events, and in some cases, significant continuous dynamics.

As a final comment, my reservations regarding the current state of PDDL2.1 should not be taken as an argument against the utility of "toy" problems and abstract languages, especially for such purposes as the IPC. However, the simplifications and abstractions employed should preserve the appropriate structure, such that research results and understanding gained in working with them can translate to real problems. This is where I believe that continued focus on classical planning models and methods will be problematic.